\documentclass[10pt,twocolumn,letterpaper]{article}

\usepackage{cvpr}
\usepackage{times}
\usepackage{epsfig}
\usepackage{graphicx}
\usepackage{amsmath}
\usepackage{amssymb}
\usepackage{boldline}
\usepackage{subfigure}
\usepackage{multirow}

\usepackage[pagebackref=true,breaklinks=true,letterpaper=true,colorlinks,bookmarks=false]{hyperref}

\cvprfinalcopy 

\def\cvprPaperID{} 

\ifcvprfinal\fi
\begin{document}

\title{GraspNet: A Large-Scale Clustered and Densely Annotated Dataset\\ for Object Grasping}

\author{Hao-Shu Fang, Chenxi Wang, Minghao Gou, Cewu Lu  \\
Shanghai Jiao Tong University\\
{\tt\small fhaoshu@gmail.com, \{wcx1997, gmh2015, lucewu\}@sjtu.edu.cn}
}

\maketitle

\begin{abstract}
Object grasping is critical for many applications, which is also a challenging computer vision problem. However, for clustered scene, current researches suffer from the problems of insufficient training data and the lacking of evaluation benchmarks. In this work, we contribute a large-scale grasp pose detection dataset with an unified evaluation system. Our dataset contains 87,040 RGBD image with over 370 million grasp poses. Meanwhile, our evaluation system directly reports whether a grasping is successful or not by analytic computation, which is able to evaluate any kind of grasp poses without exhausted labeling pose ground-truth. We conduct extensive experiments to show that our dataset and evaluation system can align well with real-world experiments. Our dataset, source code and models will be made publicly available.
\end{abstract}


\section{Introduction}
Object grasping is a fundamental problem and has many applications in industry, agriculture and service trade. The key of grasping is to detect the grasp pose given visual inputs (image or point cloud) and has drawn many attentions in computer vision community~\cite{fang2018learning,pinto2016supersizing}.

\begin{figure}[t]
    \centering
    \includegraphics[width=0.45\textwidth]{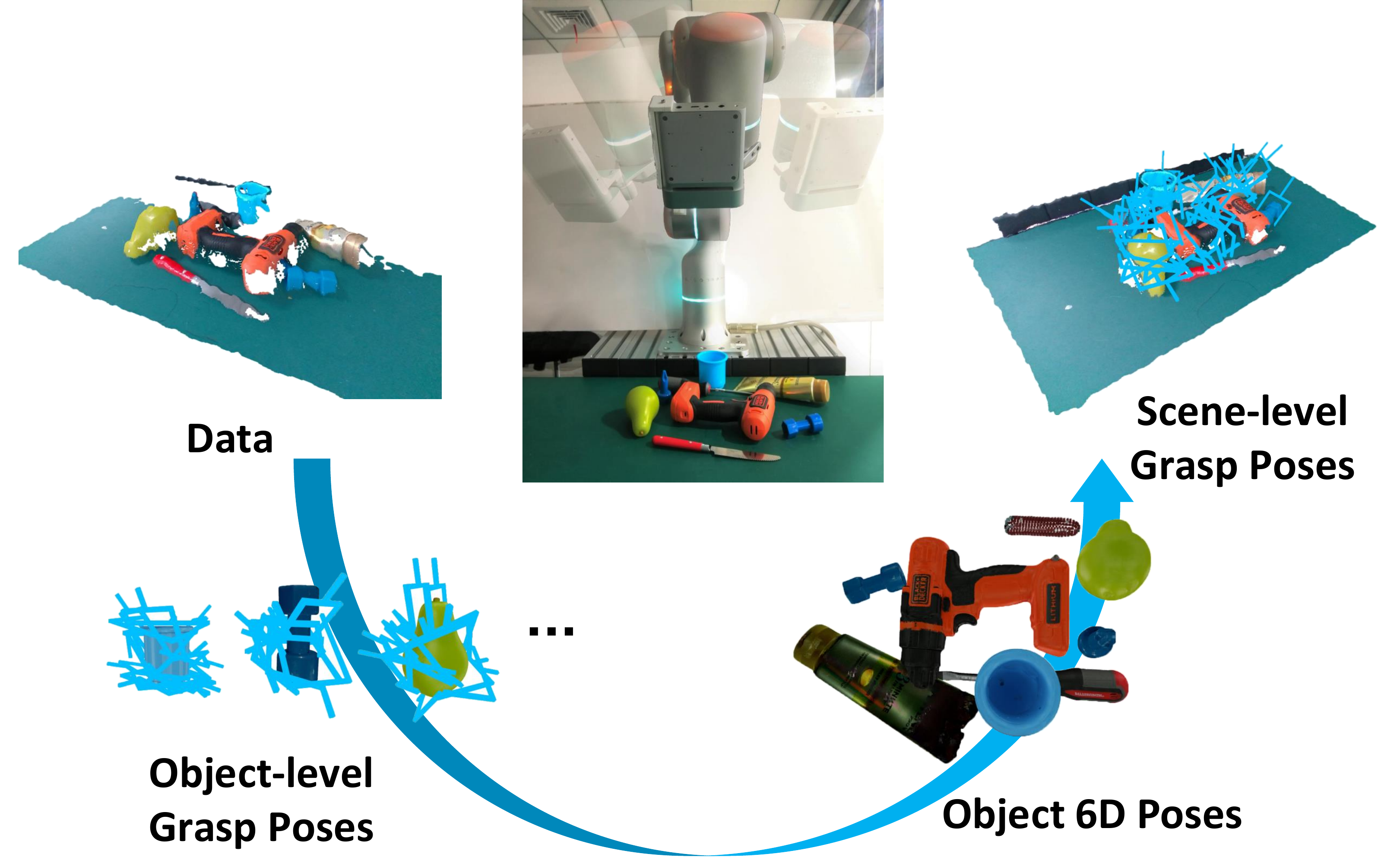}
    \caption{Our methodology for building the dataset. We collect data with real-world sensors and annotate grasp poses for every single object by analytic computation. Object 6D poses are manually annotated to project the grasp poses from object coordinate to the scene coordinate. Such methodology greatly reduces the labor of annotating grasp poses. Our dataset is both densely annotated and visually coherent with real world.}
    \label{fig:overview_of_work}

\end{figure}

Though important, there are currently two main hindrances to obtain further performance gains in this area. Firstly, the grasp pose has different representations including rectangle~\cite{redmon2015real} and 6D pose~\cite{ten2017grasp} representation and are evaluated with different metrics~\cite{jiang2011efficient, hinterstoisser2012model, ten2017grasp} correspondingly. The difference in evaluation metrics makes it difficult to compare these methods directly in an unified manner, while evaluating with real robots would dramatically increase the evaluation cost. Secondly, it is difficult to obtain large-scale high quality training data~\cite{caldera2018review}. Previous datasets annotated by human~\cite{jiang2011efficient, zhang2018roi, chu2018real} are usually small in scale and only provide sparse annotations. While obtaining training data from the simulated environment~\cite{mahler2017dex,depierre2018jacquard,yan2018learning} can generate large scale datasets, the visual domain gap between simulation and reality would inevitably degrade the performance of algorithms in real-world application.

To form a solid foundation for algorithms built upon, it is important for a dataset to i) provide data that aligns well with the visual perception from real world sensor, ii) be densely and accurately annotated with large-scale grasp pose ground-truth and iii) evaluate grasp poses with different representations in a unified manner. This is nontrivial, especially when it comes to the data annotation. Given an image or scene, it's hard for us to manually annotate endless grasp poses in continuous space. We circumvent this issue by exploring a new direction, that is, collecting data from the real world and annotating them by analytic computation in simulation, which leverages the advantages from both sides.

Specifically, inspired by previous literature~\cite{ten2017grasp}, we propose a two-step pipeline to generate tremendous grasp poses for a scene. Thanks to our automatic annotation process, we built the first large-scale in-the-wild grasp pose dataset that can serve as a base for training and evaluating grasp pose detection algorithms. Our dataset contains \textbf{87,040} RGB-D images taken from different viewpoints of over 170 clustered scenes.

For all 88 objects in our dataset, we provide accurate 3D mesh models. Each scene is densely annotated with object 6D pose and grasp pose, resulting in over \textbf{370 million} grasp poses, which is 3 orders of magnitude larger than previous datasets. Moreover, embedded with an online evaluation system, our benchmark is able to evaluate current mainstream grasping detection algorithms. Experiments also demonstrate that our benchmark can align well with real-world experiments. Fig~\ref{fig:overview_of_work} shows the methodology for building our dataset.

\section{Related Work}

In this section, we first review deep learning based grasping detection algorithms, followed by related datasets in this area.

\paragraph{Deep Learning Based Grasping Prediction Algorithms} For deep learning based grasping detection algorithms, they can be divided into three main categories. The most popular one is to detect a graspable rectangle based on RGB-D image input~\cite{jiang2011efficient, lenz2015deep, redmon2015real, guo2017hybrid, pinto2016supersizing, levine2018learning, mahler2017dex, zhang2018roi, asif2018ensemblenet, asif2018graspnet, chu2018real, morrison2018closing, mousavian20196}. Lenz~\etal~\cite{lenz2015deep} proposed a cascaded method with two networks that first prunes out unlikely grasps and then evaluates the remaining grasps with a larger network. Redmon~\etal~\cite{redmon2015real} proposed different network structure that directly regresses the grasp poses in a single step manner, which is faster and more accurate. Mahler~\etal~\cite{mahler2017dex} proposed a grasp quality CNN to predict the robustness score of grasping candidates. Zhang~\cite{zhang2018roi} and Chu~\cite{chu2018real} extended it to multi-object scenarios. The grasp poses generated by these methods are constrained in 2D plane which limits the degree of freedom of grasp poses. With the rapid development in monocular object 6D pose estimation~\cite{kehl2017ssd, xiang2017posecnn, zhao2018estimating}, some researchers~\cite{deng2019self} predicted 6D pose of the objects and project predefined grasp pose on the scene. Such methods have no limitation of grasping orientation, but require a prior knowledge about the object shape. Recently, there is a new line of researches~\cite{ten2017grasp, liang2019pointnetgpd, mousavian20196, qin2019s4g} that proposed grasping candidates on partial observed point cloud and output a classification score for each candidate using 3D CNN. Such methods have no limitation and require no prior knowledge about the objects.  Currently, these methods are evaluated in their own metrics and hard to compare to others.

\paragraph{Grasping Dataset} Cornell grasping dataset~\cite{jiang2011efficient} first proposed rectangle representation for grasping detection in images. Single object RGB-D images are provided with rectangle grasp poses.  ~\cite{chu2018real, zhang2018roi} built datasets with the same protocol but extend to multi-object scenarios.  These grasp poses are annotated by human. ~\cite{pinto2016supersizing, levine2018learning} collected annotations with real robot experiments. These data labeling methods are time consuming and require strong hardware support. To avoid such problem, some recent works explored using simulated environment~\cite{mahler2017dex,  depierre2018jacquard, yan2018learning, mousavian20196} to anotate grasp poses. They can generate a much larger scale dataset but the domain gap of visual perception is always a hindrance. Beyond rectangle based annotation, GraspSeg~\cite{asif2018graspnet} provided pixel-wise annotations for grasp-affordance segmentation and object segmentation. For 6D pose estimation, ~\cite{xiang2017posecnn} contributed a dataset with 21 objects and 70 scenes. These datasets mainly focused on a subarea of grasp pose detection. In this work, we aim to build a dataset that is  much larger in scale and diversity and covers main aspects of object grasping detection.

\begin{figure*}[t]
    \centering
    \includegraphics[width=0.95\textwidth]{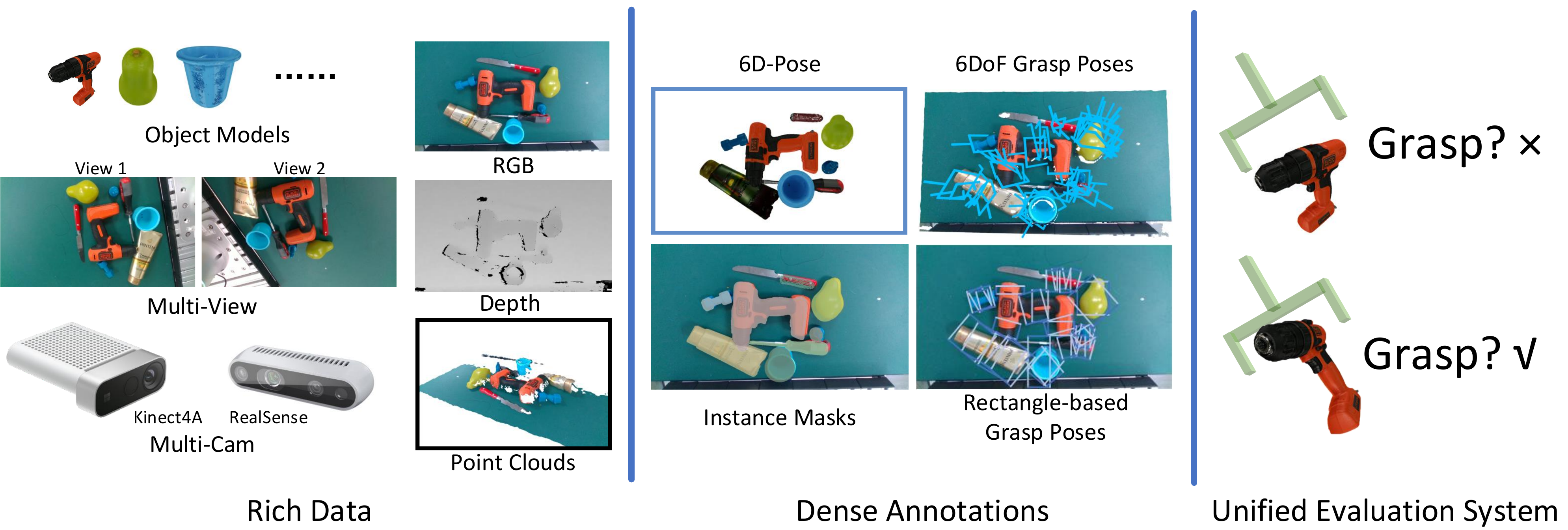}
    \caption{The key components of our dataset. RGB-D images are taken using both RealSense camera and Kinect camera from different views. The 6D pose of each object, the grasp poses, the rectangle grasp poses and the instance masks are annotated. A unified evaluation system is also provided.}
\label{fig:component_of_dataset}

\end{figure*}

\section{GraspNet Dataset}
We next describe the main features of our dataset and how we build it.
\subsection{Overview}
Previous grasping dataset either focuses on isolated object~\cite{jiang2011efficient, mahler2017dex, depierre2018jacquard, yan2018learning} or only labels one grasp per scene~\cite{pinto2016supersizing, levine2018learning}. Few datasets consider multi-object-multi-grasp setting and are small in scale~\cite{zhang2018roi, chu2018real} due to the labor of annotation. Moreover, most of the datasets adopt the rectangle based representation~\cite{jiang2011efficient} of grasp pose, which constrains the space for placing the gripper. To overcome these issues, we propose a large-scale dataset in clustered scenario with dense and rich annotations for grasp pose prediction named \emph{GraspNet}. The GraspNet dataset contains 88 daily objects with high quality 3D mesh models. The images are collected from 170 clustered scenes, each contributes 512 RGB-D images captured by two different cameras, resulting 87,040 images in total. For each image, we densely annotate 6-DoF grasp poses by analytic computation of force closure~\cite{nguyen1988constructing}. The grasp poses for each scene varies from 1,500,000 to 2,500,000, and in total our dataset contains over 370 million grasp poses. Besides, we also provide accurate object 6D pose annotations, rectangle based grasp poses, object masks and bounding boxes. Each frame is also associated with a camera pose, thus multi-view point cloud can be easily fused. Fig~\ref{fig:component_of_dataset} illustrates the key components of our dataset.

\subsection{Data Collection}
\begin{figure}
\subfigure[Camera Configuration]{
\begin{minipage}[t]{0.22\textwidth}
\centering
\includegraphics[height=3.2cm]{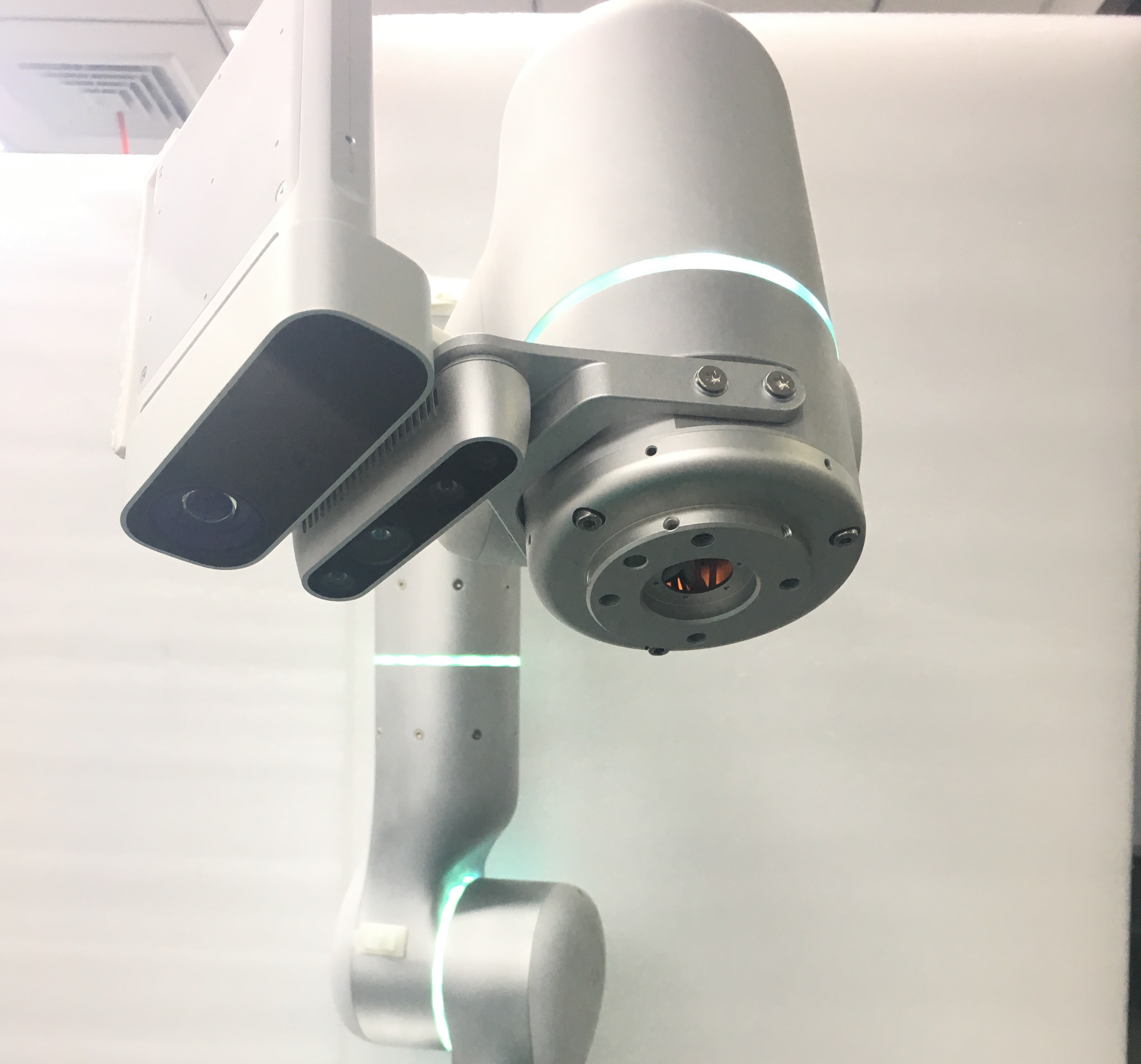}
\label{subfig:camera_configuration}
\end{minipage}%
}
\subfigure[Capturing Points]{
\begin{minipage}[t]{0.22\textwidth}
\centering
\includegraphics[height=3.2cm]{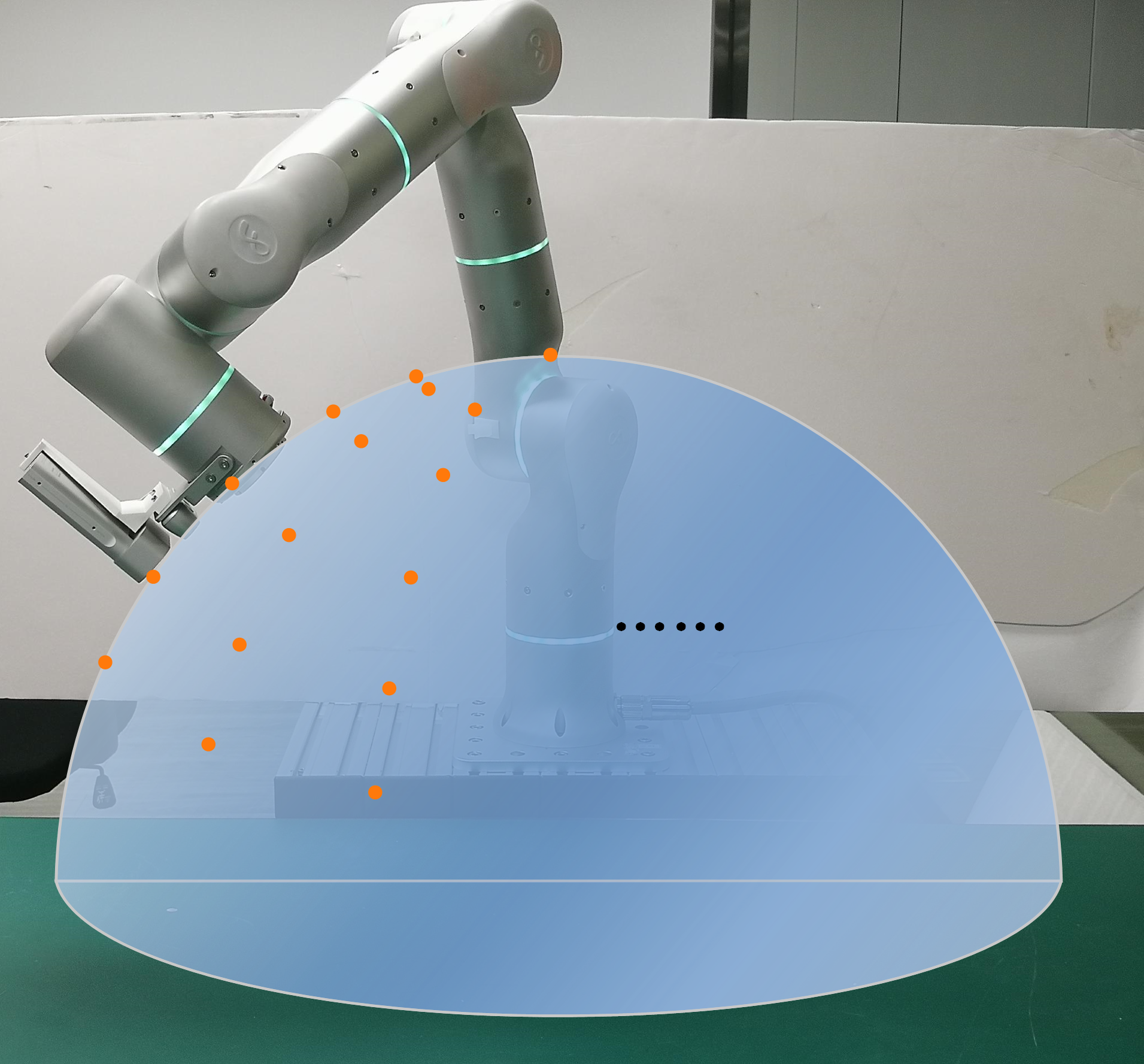}
\label{subfig:capturing_points}
\end{minipage}
}
\caption{The setting of data collection. (a) Both the RealSense and Kinect camera are fixed on the end link of a robot arm. (b) 256 data collection points are sampled from a quarter sphere.}
\label{fig:data_collection}
\vspace{-0.1in}
\end{figure}

We select 32 objects that are suitable for grasping from the YCB dataset~\cite{calli2017yale}, 13 adversarial objects from DexNet 2.0~\cite{mahler2017dex} and collect 43 objects of our own to construct our object sets. The objects vary in shape, texture, size and material. To collect data of clustered scene, we attach the cameras to a robot arm since it can repeat the trajectory precisely and help automatizing the collecting process. Hand-eye calibration is conducted before data collection to obtain accurate camera poses. Considering different quality of depth image will inevitably affect the algorithms, we adopt two popular RGB-D cameras, Intel RealSense 435 and Kinect 4 Azure, to simultaneously capture the scene and provide rich data. For each cluster scene, we randomly pick around 10 objects from our whole set and place them in a clustered manner. The robot arm then moves along a fixed trajectory that covers 256 distinct viewpoints on a quarter sphere. A synchronized image pair from both RGB-D cameras as well as their camera poses will be saved.  Fig~\ref{fig:data_collection} shows the setting for our data collection.

\subsection{Data Annotation}

\begin{table*}[]
\centering
\begin{tabular}{|l|c|c|c|c|c|c|c|c|c|}
\hline
\multicolumn{1}{|c|}{{\textbf{\footnotesize{Dataset}}} } & \footnotesize{ \textbf{\begin{tabular}[c]{@{}c@{}}Grasps\\ / scene \end{tabular}} } & \footnotesize{ \textbf{\begin{tabular}[c]{@{}c@{}}Objects\\ / scene\end{tabular}} } & \footnotesize{ \textbf{\begin{tabular}[c]{@{}c@{}}Grasp\\label\end{tabular}} } & \footnotesize{ \textbf{\begin{tabular}[c]{@{}c@{}}6D\\pose\end{tabular}} } & \footnotesize{ \textbf{\begin{tabular}[c]{@{}c@{}}Total\\ objects \end{tabular}} } & \footnotesize{ \textbf{\begin{tabular}[c]{@{}c@{}}Total\\ grasps\end{tabular}} } & \footnotesize{  \textbf{\begin{tabular}[c]{@{}c@{}}Total\\images\end{tabular}} } & \footnotesize{  \textbf{Modality} } & \footnotesize{ \textbf{\begin{tabular}[c]{@{}c@{}}Data\\source\end{tabular}}} \\ \hline
\footnotesize{Cornell~\cite{jiang2011efficient} } & \footnotesize{ $\sim$8 } & \footnotesize{ 1 } & \footnotesize{ Rect. } & \footnotesize{ No } & \footnotesize{ 240 } & \footnotesize{ 8019 } & \footnotesize{ 1035 } & \footnotesize{ RGB-D } & \footnotesize{ 1 Cam.}\\ \hline
\footnotesize{Pinto~\etal~\cite{pinto2016supersizing} } & \footnotesize{ 1 } & \footnotesize{ - } & \footnotesize{ Rect. } & \footnotesize{ No } & \footnotesize{ 150 } & \footnotesize{ 50K } & \footnotesize{ 50K } & \footnotesize{ RGB-D } & \footnotesize{ 1 Cam.} \\ \hline
\footnotesize{Levine~\etal~\cite{levine2018learning} } & \footnotesize{ 1 } & \footnotesize{ - } & \footnotesize{ Rect. } & \footnotesize{ No } & \footnotesize{ - } & \footnotesize{ 800K } & \footnotesize{ 800K } & \footnotesize{ RGB-D } & \footnotesize{ 1 Cam.} \\ \hline
\footnotesize{Mahler~\etal~\cite{mahler2017dex} } & \footnotesize{ 1 } & \footnotesize{ 1 } & \footnotesize{ Rect. } & \footnotesize{ No } & \footnotesize{ 1,500 } & \footnotesize{ 6.7M  } & \footnotesize{ 6.7M } & \footnotesize{ Depth } & \footnotesize{ Sim.} \\ \hline
\footnotesize{Jacquard~\cite{depierre2018jacquard} } & \footnotesize{ $\sim$20 } & \footnotesize{ 1 } & \footnotesize{ Rect. } & \footnotesize{ No } & \footnotesize{ 11K } & \footnotesize{ 1.1M } & \footnotesize{ 54K } & \footnotesize{ RGB-D } & \footnotesize{ Sim.} \\ \hline
\footnotesize{Zhang~\etal~\cite{zhang2018roi}} & \footnotesize{ $\sim$20 } & \footnotesize{ $\sim$3 } & \footnotesize{ Rect. } & \footnotesize{ No } & \footnotesize{ - } & \footnotesize{ 100K } & \footnotesize{ 4683 } & \footnotesize{ RGB } & \footnotesize{ 1 Cam.} \\ \hline
\footnotesize{Multi-Object~\cite{chu2018real} } & \footnotesize{ $\sim$30 } & \footnotesize{ $\sim$4 } & \footnotesize{ Rect. } & \footnotesize{ No } & \footnotesize{ - } & \footnotesize{ 2904 } & \footnotesize{ 96 } & \footnotesize{ RGB-D } & \footnotesize{ 1 Cam.} \\ \hline
\footnotesize{VR-Grasping-101*~\cite{yan2018learning} } & \footnotesize{ 100 } & \footnotesize{ 1 } & \footnotesize{ 6-DOF } & \footnotesize{ Yes  } & \footnotesize{ 101 } & \footnotesize{ 4.8M } & \footnotesize{ 10K } & \footnotesize{ RGB-D } & \footnotesize{ Sim.} \\ \hline
\footnotesize{YCB-Video~\cite{xiang2017posecnn} } & \footnotesize{ None } & \footnotesize{ $\sim$5 } & \footnotesize{ None } & \footnotesize{ Yes } & \footnotesize{ 21 } & \footnotesize{ None } & \footnotesize{ 134K } & \footnotesize{ RGB-D } & \footnotesize{ 1 Cam.} \\ \hlineB{2}
\footnotesize{\textbf{GraspNet (ours)} } & \footnotesize{ \textbf{$\sim$2.2M} } & \footnotesize{ \textbf{$\sim$10} } & \footnotesize{ \textbf{6-DOF} } & \footnotesize{ \textbf{Yes} } & \footnotesize{ \textbf{88} } & \footnotesize{ \textbf{370M} } & \footnotesize{ \textbf{87K} } & \footnotesize{ \textbf{RGB-D} } & \footnotesize{ \textbf{2 Cams.}} \\ \hline
\end{tabular}
\caption{Summary of the properties of publicly available grasp datasets. ``Rect.'', ``Cam.'' and ``Sim.'' are short for Rectangle, Camera and Simulation respectively. ``-'' denotes the number is unknown. ``*'' denotes the dataset is temporarily unavailable according to the author.}
\label{table_1}
\end{table*}
\begin{figure}
\centering
\includegraphics[width=0.5\textwidth]{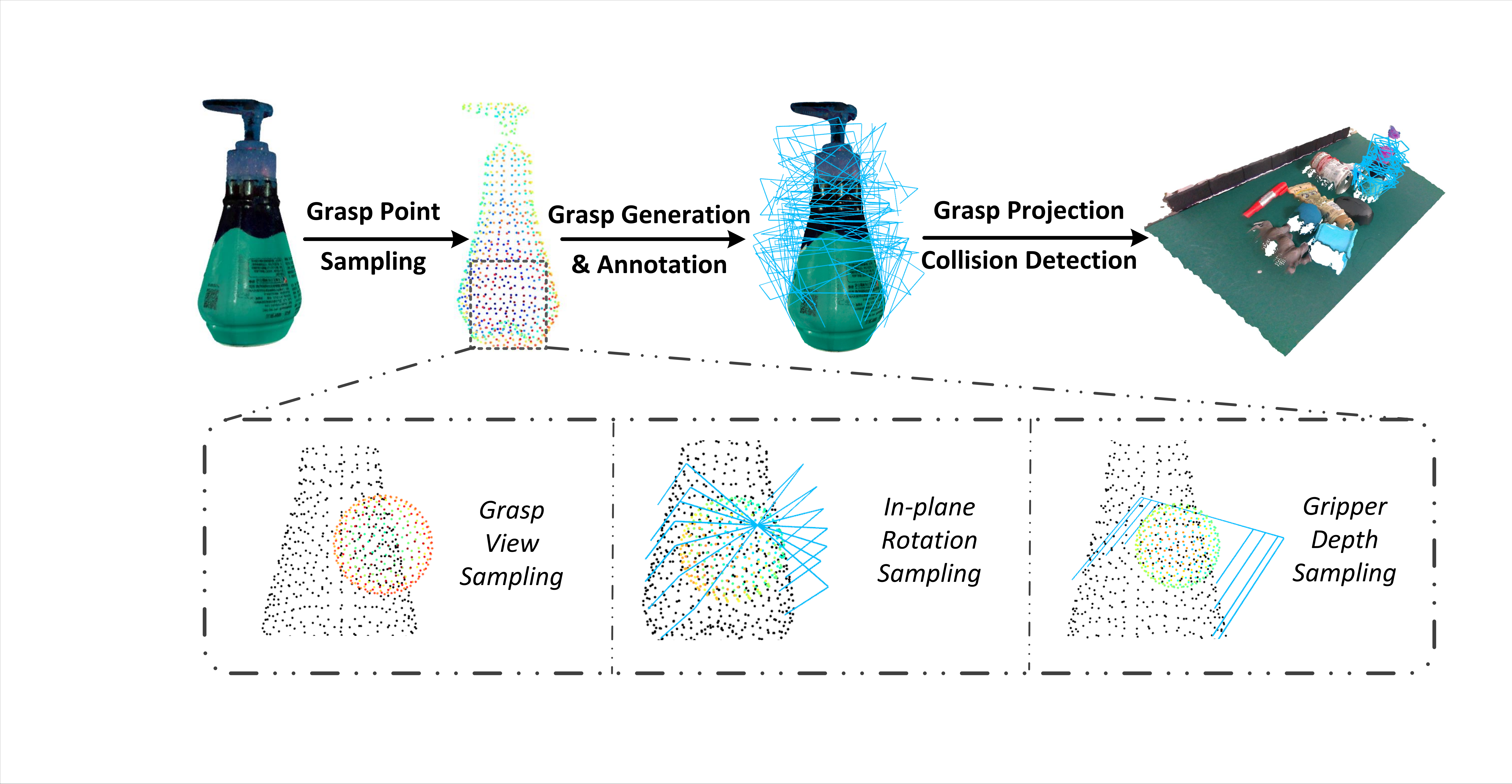}
\caption{Grasp pose annotation pipeline. The grasp point is firstly sampled from point cloud. Then the grasp view, the in-place rotation angle and the gripper depth are sampled. Grasps with high scores are adopted for each object. Finally, the grasps are projected on the scene using the 6D pose of each object. Collision detection is also conducted to avoid the collision between grasps and background or other object.}
\label{fig:annotation}
\vspace{-0.15in}
\end{figure}

\paragraph{6D Pose Annotation} With 87,040 images in total, it would be labor consuming to annotate 6D poses for each frame. Thanks to the camera poses recorded, we only need to annotate 6D poses for the first frame of each scene. The 6D poses will then be propagated to the remaining frames by:
\begin{equation}
    \mathbf{P}_i^j = \mathbf{cam}_i^{-1}\mathbf{cam}_0\mathbf{P}_0^j,
\end{equation}
where $\mathbf{P}_i^j$ is the 6D pose of object $j$ at frame $i$ and $\mathbf{cam}_i$ is the camera pose of frame $i$. Followed by ICP refinement and human checking, we can annotate 6D poses accurately in an efficient manner. Object masks and bounding boxes are also obtained by projecting objects onto the images using 6D poses.

\paragraph{Grasp Pose Annotation} Different from labels in common vision tasks, grasp poses distribute in a large and continuous search space, which brings infinite annotations. Annotating each scene manually would be dramatically labor expensive. Considering all the objects are known, we propose a two stage automated pipeline for grasp pose annotation, which is illustrated in Fig. \ref{fig:annotation}.

First, grasp poses are sampled and annotated for each single object. To achieve that, high quality mesh models are downsampled such that the sampled points (called grasp points) are uniformly distributed in voxel space. For each grasp point, we sample $V$ views uniformly distributed in a spherical space. Grasp candidates are searched in a two dimensional grid $D\times A$, where $D$ is the set of gripper depths and $A$ is the set of in-plane rotation angles. Gripper width is determined accordingly such that no empty grasp or collision occurs. Each grasp candidate will be assigned a confidence score based on the mesh model.

We adopt an analytic computation method to grade each grasp. The force-closure metric~\cite{nguyen1988constructing, ten2017grasp} has been proved effective in grasp evaluation: given a grasp pose, the associated object and a friction coefficient $\mu$, force-closure metric outputs a binary label indicating whether the grasp is antipodal under that coefficient. The result is computed based on physical rules, which is robust. Here we adopt an improved metric described in~\cite{liang2019pointnetgpd}. With $\Delta\mu=0.1$ as interval, we increase $\mu$ gradually from $0.1$ to $1$ step by step until the grasp is antipodal. The grasp with lower friction coefficient $\mu$ has more probability of success. Thus we define our score $s$ as:
\begin{equation}
s = 1.1 - \mu,
\vspace{-0.05in}
\label{eqn:confidence}
\end{equation}
such that $s$ lies in $(0,1]$.

Second, for each scene, we project these grasps to the corresponding objects based on the annotated 6D object poses:
\begin{equation}
\vspace{-0.05in}
\begin{split}
& \mathbf{P}^i = \mathbf{cam_0}\mathbf{P}_0^i,\\
& \mathbb{G}^i_{(w)} = \mathbf{P}^i\cdot \mathbb{G}^i_{(o)},
\label{eqn:transform}
\end{split}
\end{equation}
where $\mathbf{P}^i$ is the 6D pose of the $i$-th object in the world frame, $\mathbb{G}^i_{(o)}$ is a set of grasp poses in the object frame and $\mathbb{G}^i_{(w)}$ contains the corresponding poses in the world frame. Besides, collision check is performed to avoid invalid grasps. After these two steps we can generate densely distributed grasp set $\mathbb{G}_{(w)}$ for each scene. According to statistics, the ratio of positive and negative labels in our dataset is around 1:2. We conduct real world experiment in Sec.~\ref{sec:exp} using our robot arm and verify that our generated grasp poses can align well with real world grasping.

\subsection{Evaluation}
\paragraph{Dataset Split} For our 170 scenes, we use 100 for training and 70 for testing. Specifically, we further divide our test sets into 3 categories: 30 scenes with seen objects, 30 with unseen but similar objects and 10 for novel objects. We hope that such setting can better evaluate the generalization ability of different methods.

\paragraph{New Metrics}
To evaluate the prediction performance of grasp pose, previous methods adopt the rectangle metric that consider a grasp as correct if: i) the rotation error is less than 30$^{\circ}$ and ii) the rectangle IOU is larger than 0.25.

There are several drawbacks of such metric. Firstly, it can only evaluate rectangle representation of grasp pose. Secondly, the error tolerance is set rather high since the groundtruth annotations are not exhaustive. It might overestimate the performance of grasping algorithm. Currently, the Cornell dataset~\cite{jiang2011efficient} has achieved over 99\% accuracy. In this work, we adopt an online evaluation algorithm to evaluate the grasp accuracy.

We first illustrate how we classify whether a single grasp pose is true positive. For each predicted grasp pose $\mathbf{\hat{P}}_i$, we associate it with the target object by checking the point cloud inside the gripper. Then, similar to the process of generating grasp annotation, we can get a binary label for each grasp pose by force-closure metric, given different $\mu$.

For clustered scene, grasp pose prediction algorithms are expected to predict multiple grasps. Since for grasping, we usually conduct execution after the prediction, the percentage of true positive is more important. Thus, we adopt \emph{Precision@k} as our evaluation metric. $AP$ denotes the average \emph{Precision@k} for $k$ ranges from 1 to 50. Similar to COCO~\cite{coco}, we report $AP$ at different $\mu$. Specifically, we denote \textbf{AP} for $AP$ from $\mu=0.1$ to $\mu=0.5$, with $\Delta\mu= 0.1$ as interval.

To avoid dominated by similar grasp poses or grasp poses from single object, we run a pose-NMS before evaluation.
For two grasps $\mathbf{G}_1$ and $\mathbf{G}_2$, we define grasp pose distance $D(\mathbf{G}_1, \mathbf{G}_2)$ as a tuple:
\begin{equation}
    D(\mathbf{G}_1, \mathbf{G}_2) = (d_t(\mathbf{G}_1, \mathbf{G}_2), d_{\alpha}(\mathbf{G}_1, \mathbf{G}_2)),
\end{equation}
where $d_t(\mathbf{G}_1, \mathbf{G}_2)$ and $d_{\alpha}(\mathbf{G}_1, \mathbf{G}_2))$ denote translation distance and rotation distance of two grasps respectively. Let a grasp pose $\mathbf{G}$ be denoted by a translation vector $\mathbf{t}$ and a rotation matrix $\mathbf{R}$, then $d_t(\cdot)$ an $d_{\alpha}(\cdot)$ is defined as:
\begin{equation}
    \begin{split}
        d_t(\mathbf{G}_1, \mathbf{G}_2) &= ||\mathbf{t}_1 - \mathbf{t}_2||,\\
        d_{\alpha}(\mathbf{G}_1, \mathbf{G}_2) &= \arccos \frac{1}{2} (\mathrm{tr}(R_1\cdot R_2^\mathrm{T}) - 1),
    \end{split}
\end{equation}
where $\mathrm{tr}(\mathbf{M})$ denotes the trace of matrix $\mathbf{M}$.

Since translation and rotation are not in the same metric space, we define the NMS threshold as a tuple too. Let $TH = (th_d, th_{\alpha})$, we say $D(\mathbf{G}_1, \mathbf{G}_2) < TH$ if and only if
\begin{equation}
    d_t(\mathbf{G}_1, \mathbf{G}_2) < th_d,\quad d_{\alpha}(\mathbf{G}_1, \mathbf{G}_2) < th_{\alpha}.
\end{equation}

Based on the tuple metric, two grasps are merged when their distance is lower than $TH$. Meanwhile, only the top $K$ grasps from each object are considered according to confidence scores and other grasps are omitted. In evaluation, we set $th_d = 1$ cm, $th_{\alpha} = 5$ degree and $K = 10$.

\subsection{Discussion}\label{subsec:dataset_discussion}

We compare our datasets with other publicly available grasp datasets. Table~\ref{table_1} summaries the main differences at several aspects. We can see that our dataset is much larger in scale and diversity. With our two-step annotation pipeline, we are able to collect real images with dense annotations, which leverages the advantages from both sides.

For grasp pose evaluation, due to the continuity in grasping space, there are in fact infinite feasible grasp poses. The previous method that pre-computed ground truth for evaluating grasping, no matter collected by human annotation~\cite{jiang2011efficient} or simulation~\cite{depierre2018jacquard}, cannot cover all feasible solution. In contrast, we do not pre-compute labels for the test set, but directly evaluate them by calculating the quality score using force closure metric~\cite{nguyen1988constructing}. Such evaluation method does not assume the representation of the grasp pose, thus is general in practice. Related APIs will be made publicly available to facilitate the research in this area.

\section{Experiments}\label{sec:exp}
In this section, we conduct robotic experiments to demonstrate that our ground-truth annotations can align well with real-world grasping.
\subsection{Ground-Truth Evaluation}
\begin{figure}[t]
\centering
\subfigure[6D Object Pose]{
\begin{minipage}[t]{0.47\linewidth}
\centering
\includegraphics[height=3cm]{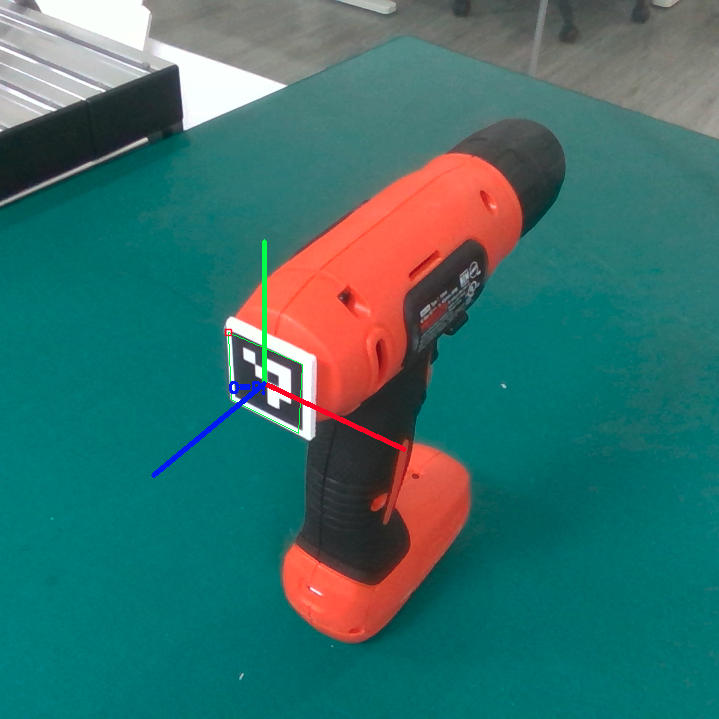}
\label{subfig:object_6d_pose}
\end{minipage}%
}
\subfigure[grasp poses]{
\begin{minipage}[t]{0.47\linewidth}
\centering
\includegraphics[height=3cm]{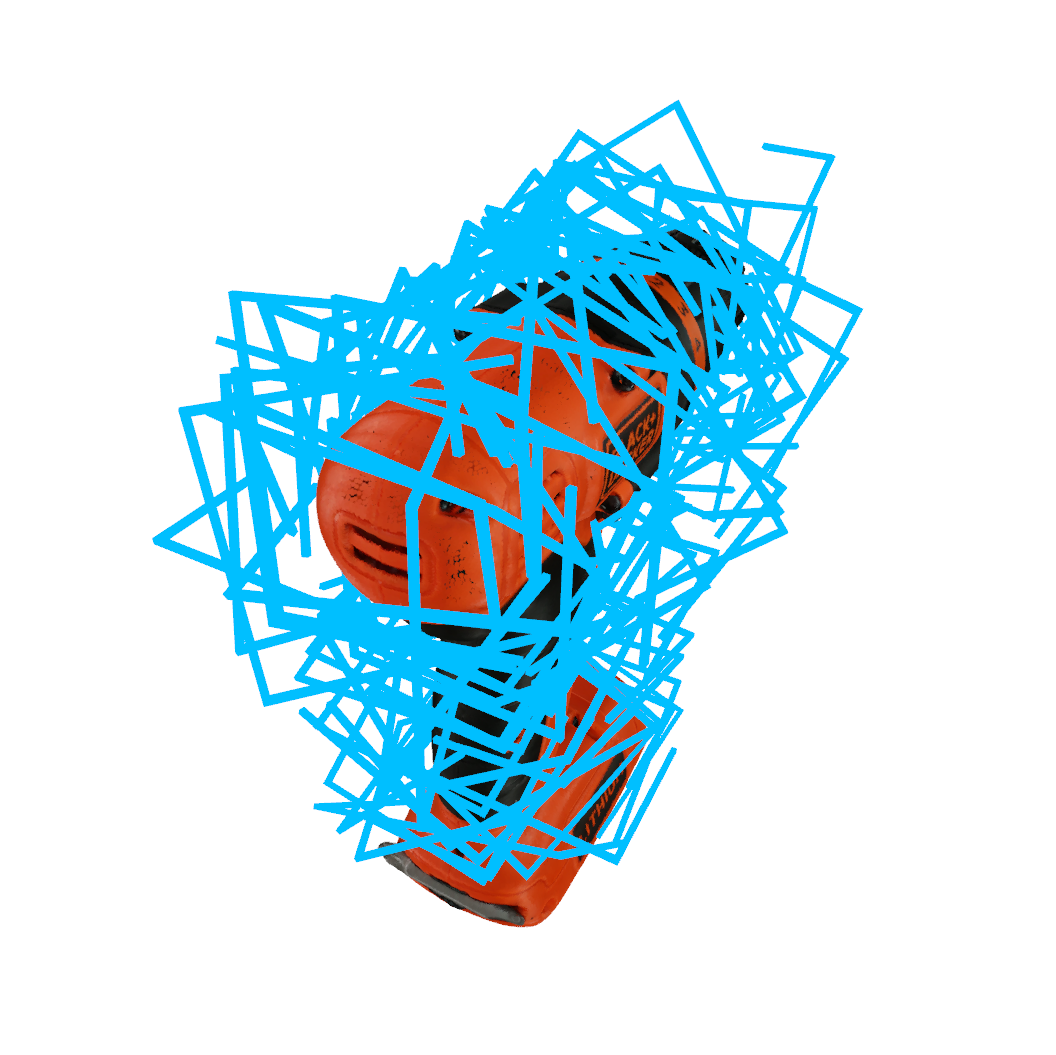}
\label{subfig:grasping_poses}
\end{minipage}
}
\caption{(a) Visualization of the detected 6D pose the ArUco marker. (b) Object 6D pose and grasp poses inferred from the marker pose.}
\label{fig:6d_and_grasping_pose}
\end{figure}

To evaluate the quality of our generated grasp poses, we set up a real robotic experiment. Since we need to project grasp poses to the camera frame using objects' 6D poses, we paste ArUco code on the objects and only label their 6D poses once to avoid tedious annotation process. Fig~\ref{fig:6d_and_grasping_pose} illustrates our grasp pose projection process.

We pick 10 objects from our object set and execute grasp poses that has different scores. For each setting we randomly choose 100 grasp poses. For robot arm we adopt a Flexiv Rizon arm and for camera we use the Intel RealSense 435. Table~\ref{table_real} summarizes the success rate of grasping. We can see that for grasp poses with high score, the success rate can achieve 0.96 in average. Meanwhile, the success rate is pretty low for grasp poses with $s=0.1$. It indicates that our generated grasp poses are well aligned with real world grasping.

\begin{table}[]
\centering
\begin{tabular}{|c|c|c|c|c|c|c|c|}
\hline

\scriptsize{\textbf{Object}} & \scriptsize{s=1} & \scriptsize{s=0.5} & \scriptsize{s=0.1} & \scriptsize{\textbf{Object}} & \scriptsize{s=1} & \scriptsize{s=0.5} & \scriptsize{s=0.1} \\ \hline
\scriptsize{Banana} & \scriptsize{98\%} & \scriptsize{67\%} & \scriptsize{21\%} & \scriptsize{Apple} & \scriptsize{97\%} & \scriptsize{65\%} & \scriptsize{16\%} \\ \hline
\scriptsize{Peeler} & \scriptsize{95\%} & \scriptsize{59\%} & \scriptsize{9\%} & \scriptsize{Dragon} & \scriptsize{96\%} & \scriptsize{60\%} & \scriptsize{9\%} \\ \hline
\scriptsize{Mug} & \scriptsize{96\%} & \scriptsize{62\%} & \scriptsize{12\%} & \scriptsize{Camel} & \scriptsize{93\%} & \scriptsize{67\%} & \scriptsize{23\%} \\ \hline
\scriptsize{Scissors} & \scriptsize{89\%} & \scriptsize{61\%} & \scriptsize{5\%} & \scriptsize{Power Drill} & \scriptsize{96\%} & \scriptsize{61\%} & \scriptsize{14\%} \\ \hline
\scriptsize{Lion} & \scriptsize{98\%} & \scriptsize{68\%} & \scriptsize{16\%} & \scriptsize{Black Mouse} & \scriptsize{98\%} & \scriptsize{64\%} & \scriptsize{13\%} \\ \hline
\end{tabular}
\caption{Summary of real world success rate of grasping given different grasp score.}
\label{table_real}

\end{table}

\section{Conclusion}
In this paper we built a large-scale dataset for clustered scene object grasping. Our dataset is orders of magnitude larger than previous grasping datasets and diverse in objects, scenes and data sources. It consists of images taken by real world sensor and has rich and dense annotations. Meanwhile, a unified evaluation system is also proposed to promote the development of this area. We demonstrated that our dataset and evaluation system align well with real world grasping. In the future, we will also extend our dataset to multi-finger gripper and vacuum-based end effectors. Our code and dataset will be released.
{\small
\bibliographystyle{ieee_fullname}
\bibliography{egbib}
}

\end{document}